\pdfoutput=1

\documentclass[11pt]{article}

\usepackage[]{acl}

\usepackage{times}
\usepackage{latexsym}

\usepackage[T1]{fontenc}

\usepackage[utf8]{inputenc}

\usepackage{microtype}

\usepackage{url}            
\usepackage{booktabs}       
\usepackage{amsfonts}       
\usepackage{nicefrac}       
\usepackage{microtype}      

\usepackage{multicol}
\usepackage{tabularx}
\usepackage{graphicx}
\usepackage{threeparttable}
\usepackage{adjustbox}

\usepackage{amsmath}
\usepackage{amssymb}
\usepackage{pifont}
\usepackage{wasysym}
\usepackage{enumitem}

\DeclareMathOperator*{\argmax}{arg\,max}

\title{NER-MQMRC: Formulating \underline{N}amed \underline{E}ntity \underline{R}ecognition as \underline{M}ulti \underline{Q}uestion \underline{M}achine \underline{R}eading \underline{C}omprehension}

\author{Anubhav Shrimal\textsuperscript{1} \And Avi Jain\textsuperscript{2}\thanks{~~work done as part of Retail Business Services, Amazon} \And Kartik Mehta\textsuperscript{3}\thanks{~~work done as part of India Machine Learning, Amazon} \And Promod Yenigalla\textsuperscript{1} \\\AND
    \\
    \textsuperscript{1}Retail Business Services, Amazon \\
    \textsuperscript{2}Product Graph, Amazon \\
    \textsuperscript{3}Amazon Search, Amazon \\
    \texttt{\{shrimaa, jainav, kartim, promy\}@amazon.com}\\
}

\begin{document}
\maketitle
\begin{abstract}
NER has been traditionally formulated as a sequence labeling task. However, there has been recent trend in posing NER as a machine reading comprehension task \citep{google-kdd-2020, Xue2020CoarsetoFinePF}, where entity name (or other information) is considered as a question, text as the context and entity value in text as answer snippet. These works consider MRC based on a single question (entity) at a time. We propose posing NER as a multi-question MRC task, where multiple questions (one question per entity) are considered at the same time for a single text. We propose a novel BERT-based multi-question MRC (NER-MQMRC) architecture for this formulation. NER-MQMRC architecture considers all entities as input to BERT for learning token embeddings with self-attention and leverages BERT-based entity representation for further improving these token embeddings for NER task. Evaluation on three NER datasets show that our proposed architecture leads to average 2.5 times faster training and 2.3 times faster inference as compared to NER-SQMRC framework based models by considering all entities together in a single pass. Further, we show that our model performance does not degrade compared to single-question based MRC (NER-SQMRC)~\citep{devlin-bert} leading to F1 gain of +0.41\%, +0.32\% and +0.27\% for AE-Pub, Ecommerce5PT and Twitter datasets respectively. We propose this architecture primarily to solve large scale e-commerce attribute (or entity) extraction from unstructured text of a magnitude of 50k+ attributes to be extracted on a scalable production environment with high performance and optimised training and inference runtimes.

\end{abstract}

\section{Introduction}
Named Entity Recognition (NER) is the task of locating and classifying entities mentioned in unstructured text into predefined categories such as names of people, organizations and locations. It is a crucial component of many applications, such as web search, relation extraction~\citep{Bowen2019BeyondWA} and e-commerce attribute extraction~\citep{opentag, Mehta2021LaTeXNumericLT}. Traditionally, NER has been posed as a sequence labeling task~\citep{Ma2016EndtoendSL, opentag, devlin-bert} where each token is assigned a single tag class. We term these sequence labeling approaches as NER-SL. Recently, there has been interest in posing NER as a machine reading comprehension task \citep{google-kdd-2020, Xue2020CoarsetoFinePF, xu-etal-2019-scaling}. Specifically, NER is posed as a question answering problem, where text is considered context, entity name (or some variant) is considered question and entity value mentioned in text is considered as answer snippet. We term these approaches as Single Question Machine Reading Comprehension (NER-SQMRC) as they involve asking a single question (or entity) at a time. We argue that both NER-SL and NER-SQMRC have their merits and demerits, e.g. NER-SQMRC incorporates entity name for better representation and can be easily extended to new entities without re-training and NER-SL requires single scoring pass for extracting all entities from a given text. We pose NER as a multi-question MRC problem, where multiple questions (one question per entity) are asked at the same time and propose a novel architecture (NER-MQMRC) for this formulation. We summarize the merits and demerits of these three formulations in Table \ref{tab:architecture-comparison} considering below factors:
\begin{itemize}[noitemsep]
\item \textbf{Entity scaling}: Ability to scale for new entities without retraining.
\item \textbf{Multi-entity scoring}: Ability to extract all entities from a given text in a single forward pass.
\item \textbf{Faster runtime}: Extracting multiple entities together in a single pass leads to faster training and inference as compared to considering 
single entity in a pass.
\item \textbf{Using entity information}: Leveraging entity information (such as entity name) for learning better representations.
\end{itemize}

\begin{table}[h]
\centering
\begin{adjustbox}{max width=0.43\textwidth}
\begin{tabular}{lccc}
\toprule
Property & NER-SL & NER-SQMRC & NER-MQMRC \\
\midrule
Entity scaling & \ding{55} & \ding{52} & \ding{52} \\
Multi-entity Scoring & \ding{52} & \ding{55} & \ding{52} \\
Faster runtime & \ding{52} & \ding{55} & \ding{52} \\
Entity information & \ding{55} & \ding{52} & \ding{52} \\
\bottomrule
\end{tabular}
\end{adjustbox}
\caption{Comparing different attribute extraction approaches based on various factors.}
\label{tab:architecture-comparison}
\end{table}

As summarized in Table \ref{tab:architecture-comparison}, our proposed NER-MQMRC architecture combines the best of NER-SL and NER-SQMRC. NER-MQMRC considers extraction of multiple entities based on multiple questions on same text, and is novel in three ways - 
1) Token representations are learnt to incorporate information of all the entities, unlike using single entity as in~\citep{google-kdd-2020, Xue2020CoarsetoFinePF}.
2) We introduce leveraging BERT-based entity representations for further improving token representations for NER task.
3) Our architecture leads to faster training and inference. E.g. scoring of five entities can be done using a single forward pass with our NER-MQMRC as compared to five passes required earlier with NER-SQMRC based models~\citep{devlin-bert, google-kdd-2020, Xue2020CoarsetoFinePF}. Experiments on three NER datasets establish the effectiveness of NER-MQMRC architecture. NER-MQMRC achieves 2.5x faster training and 2.3x faster inference as compared to single question based MRC (NER-SQMRC) framework based models by considering multiple entities together in training and inference. Further, we show performance boost over SOTA NER-SQMRC~\citep{devlin-bert}, obtaining  +0.41\%, +0.32\% and +0.27\% F1 improvements for AE-Pub, Ecommerce5PT and Twitter datasets respectively.
Rest of the paper is organized as follows. We describe our proposed NER-MQMRC architecture in Section~\ref{approach}. We discuss our experimental setup in Section~\ref{experiments} followed by results in Section~\ref{results}. We discuss the industry impact of our work in Section~\ref{industry_impact} and summarize the paper in Section~\ref{conclusion}.

\section{NER as a Multi-Question MRC task}
\label{approach}

\subsection{Problem definition and dataset construction}
\label{problem_def_dataset}
Given an input sequence $X = \{x_{1}, x_{2}, ..., x_{n}\}$, where $n$ denotes the length of the sequence, the objective in NER task is to find and label tokens in $X$ that represent entity $y~\in~Y$, where $Y$ is a predefined list of all possible entities (e.g., BRAND, COLOR, etc). Under the NER-SQMRC framework, the model is given a question $q_{i}$ asking about $i^{th}$ entity and the model has to extract a text span $x_{start_{i},end_{i}}$ from $X$ which are tokens corresponding to the $i^{th}$ entity. For NER-MQMRC framework, the model is given a list of $k$ questions $Q = \{q_{1}, q_{2}, ..., q_{k}\}$ and the model has to extract the text spans $\{(x_{start_{1},end_{1}}), (x_{start_{2},end_{2}}), ..., (x_{start_{k},end_{k}})\}$ from $X$ corresponding to each of the $k$ entities. We use BERT for Question Answering~\citep{devlin-bert} as our NER-SQMRC baseline implementation~(refer Appendix~\ref{ner_sqmrc_baseline}).

\begin{figure}[ht]
\centering
\includegraphics[width=0.40\textwidth]{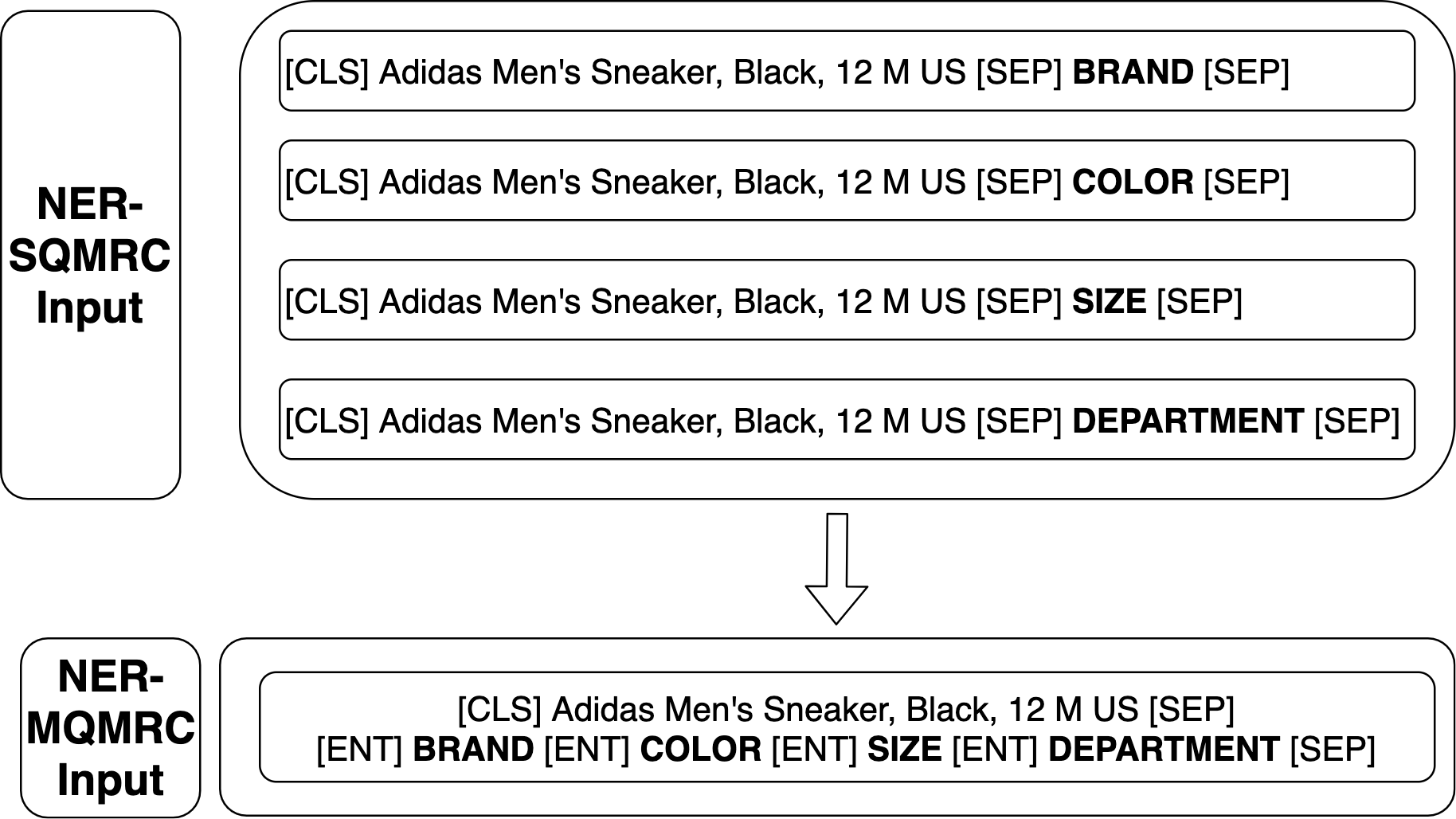}
\caption{Data Input format for NER-SQMRC and NER-MQMRC model architectures.}
\label{fig:input_format}
\end{figure}

Figure~\ref{fig:input_format} shows data input format for both NER-SQMRC and NER-MQMRC. Similar to conventional Question Answering, training data for NER-SQMRC consists of (text, single-entity-question, entity spans from text) triplets. For a dataset with $k$ entities, training data consists of $k$ samples for each text, each sample having question for one entity. However, for NER-MQMRC, training data consists of a single sample for each text, having $k$ questions (one question per entity). Hence, NER-SQMRC formulation requires dealing with larger size training data ($k$ times more samples) with same information as compared to NER-SL and NER-MQMRC. Similarly, during inference, NER-SQMRC requires performing $k$ evaluations for the same text to get text span for each entity, whereas NER-MQMRC requires only a single evaluation for all entities.

\subsection{Model Details}
\begin{figure}[ht]
\centering{\includegraphics[width=0.48\textwidth]{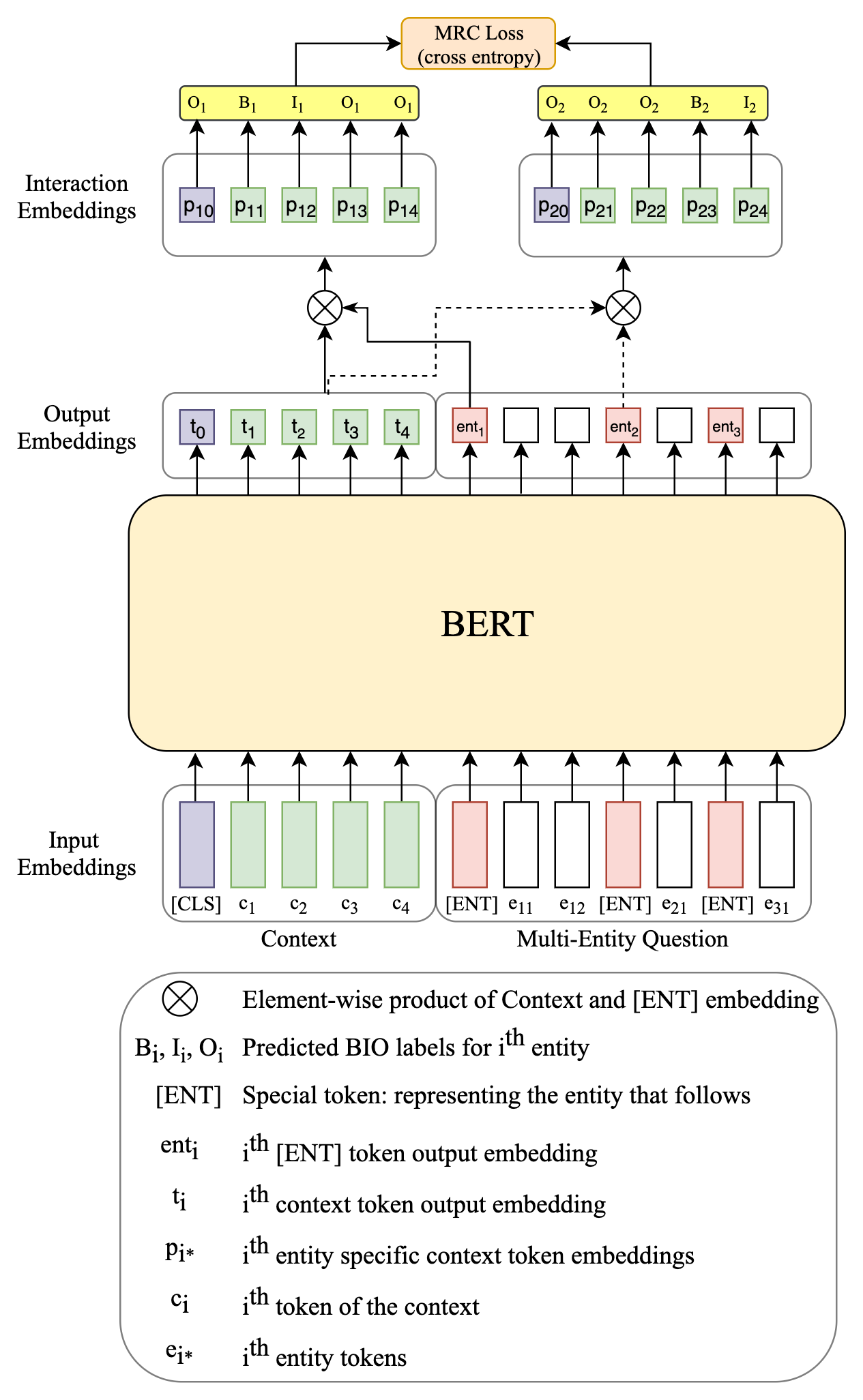}}
\caption{NER-MQMRC model architecture.}
\label{fig:model_arch}
\end{figure}

Figure~\ref{fig:model_arch} shows our proposed NER-MQMRC architecture. We build NER-MQMRC on top of BERT architecture~\citep{devlin-bert} by customizing BERT input and modifying the output layer as described in this section.

\subsubsection{NER-MQMRC input}
\label{bert-input}
BERT has been trained to take a pair of sentences separated by a special token \textit{[SEP]} as input, and use $E{_A}$ and $E{_B}$ segment embeddings respectively for tokens of each sentence. For NER-MQMRC, we concatenate the input text and questions of all entities separated by \textit{[SEP]} (refer Figure~\ref{fig:input_format}). Questions of each entity are further separated by a special token \textit{[ENT]}, which we add to the BERT vocabulary. We use $E{_A}$ segment embeddings for input text and $E{_B}$ segment embeddings for all question tokens. Output embedding learned corresponding to each \textit{[ENT]} token is considered as embedding representation for the entity adjacent to that \textit{[ENT]} token.

\subsubsection{Entity specific representation and span selection}
\label{attention-layer}
As discussed, the $i^{th}$ \textit{[ENT]} token output embedding ($ent_i$) represents the $i^{th}$ entity in the question. We hypothesize that using $ent_i$ to attend to the context token's output embeddings, $T = \{t_1, t_2, ..., t_n\}$, will help the model find the answer span for entity $i$. We use entity embeddings to transform the common context representations ($T$) to entity specific token representations. We consider extraction of each entity as a separate task and use element-wise product of token and entity embeddings to obtain entity specific representations for each token (refer Figure~\ref{fig:model_arch}). More formally, we perform an element-wise product of token embeddings $T$ with $ent_i$ to get $i^{th}$ entity specific token representations $P_i = \{p_{i1}, p_{i2}, ..., p_{in}\}$. 

These entity specific representations are then fed into a separate token-level dense layer, $W_{bio}$, to get the BIO format label prediction for each token w.r.t. the entity as shown in equation~\ref{eq:bio}, where $t_{j}$ represents embedding for $j^{th}$ token and $ent_{i}$ represents embedding for $i^{th}$ entity. Examples with no entity mention are modelled by setting the label for \textit{[CLS]} token as $B$ tag for that entity. For each token and entity pair, loss is calculated using cross entropy loss ($L^{ce}$) between predicted and actual label. For each sample, total loss, $L_{total}$ (refer equation~\ref{eq:loss}), is average (with equal weightage) of loss for all $k$ entities and  $n$ tokens pairs.


\begin{equation} 
\label{eq:bio}
label_{j} = \argmax (softmax(W_{bio}(t_j \odot ent_i)))
\end{equation}

\begin{equation} \label{eq:loss}
L_{total} = \frac{1}{k \cdot n} \sum_{i=1}^{k} \sum_{j=1}^{n} L^{ce}_{i,j}
\end{equation}

\subsection{Discussion}
Our proposed formulation is generic and can be used with other pre-trained architectures (such as XLNET, RoBERTa) instead of BERT for feature extraction. In recent years, there has been incremental advancements to the MRC framework such as the use of knowledge distillation loss as a regularizer and no-answer loss~\citep{google-kdd-2020} to achieve better performance than~\citep{devlin-bert}; NER-MQMRC framework can also easily integrate such ideas to get better performance and we keep this to be explored as a future work since in this paper we want to show the effectiveness of NER-MQMRC framework over NER-SQMRC framework with similar setup for both the frameworks. We use BIO label prediction to allow multiple value predictions for an entity from the text, though we also experimented with single (start, end) span index prediction as output labels similar to~\citep{google-kdd-2020} but has the limitation of predicting only a single answer span.

\section{Experimental Setup}
\label{experiments}

\subsection{Datasets}
\label{sec:dataset}

\begin{table}[]
\centering
\begin{adjustbox}{max width=0.48\textwidth}
\begin{tabular}{@{}lcccccc@{}}
\toprule
                & \multicolumn{3}{c}{Train Data}    & \multicolumn{3}{c}{Test Data}   \\
Dataset         & SQMRC   & MQMRC   & Reduction(\%)     & SQMRC  & MQMRC  & Reduction(\%)     \\ \midrule
Ecommerce5PT       & 981,076 & 290,698 & \textbf{70.37} & 32,062 & 4,967  & \textbf{84.51} \\
AE-Pub          & 88,460  & 39,888  & \textbf{54.91} & 22,005 & 17,393 & \textbf{20.96} \\
Twitter         & 11,997  & 3,999   & \textbf{66.67} & 9,768  & 3,256  & \textbf{66.67} \\ \bottomrule
\end{tabular}
\end{adjustbox}
\caption{Reduction in dataset size due to single-entity to multi-entity question transformation.}
\label{tab:dataset_rows_stats}
\end{table}

Experiments were performed on three NER datasets described below.

\noindent \textbf{AE-Pub}~\citep{xu-etal-2019-scaling}
is a dataset for E-commerce attribute extraction collected from AliExpress Sports \& Entertainment category. This dataset is designed to pose E-commerce attribute extraction as a question answering problem and contains over 110k triplets (text, attribute, value) and 2.7k unique attributes. Even though the number of attributes is large, any given text in the dataset has no more than 13 attributes. Train and test dataset is created in an automated manner using distant supervision.

\noindent \textbf{Ecommerce5PT}
is a 33 attributes (size, material, color, etc.) dataset extracted from five different product types from Amazon catalogue. The train data is constructed in a similar way as AE-Pub using distant supervision. The train data quality is improved using automated gazetteer and matching heuristics (refer Appendix~\ref{ecommerce5pt_data}). Unlike AE-pub, test data is constructed with manual audit, thus leading to better quality test data.


\noindent \textbf{Twitter}~\citep{Zhang2018AdaptiveCN}
is an English NER dataset based on tweets. We use the setup similar to~\citep{Xue2020CoarsetoFinePF}, using textual information queries (refer Appendix~\ref{queries}) and making entity detection on PER, LOC and ORG.

\subsubsection{Datasets transformation}
As discussed earlier, NER-MQMRC leads to reduced train and test data size as compared to NER-SQMRC (Table~\ref{tab:dataset_rows_stats}). We observe a median of 3, 2 and 3 entities per question in training data of Ecommerce5PT, AE-Pub and Twitter datasets respectively, leading to similar data reduction for NER-MQMRC training. Appendix~\ref{entities_per_que} elaborates on the distribution of entities per question for NER-MQMRC for each of these datasets. For fair comparison, one should use all entities of a sample while evaluating NER-SL, NER-SQMRC and NER-MQMRC approaches. We use this setup for Ecommerce5PT and Twitter datasets. However, AE-Pub dataset contains only few entities of each sample. We follow setup used in~\citep{xu-etal-2019-scaling} for AE-Pub evaluation.

\subsection{Experiments}
In this section we detail the various experiments to evaluate our proposed solution, NER-MQMRC, on aspects such as operational performance (training and inference runtime), NER task, limited data setting (few shot) and NER-MQMRC model specific analysis.

\noindent \textbf{Training and Inference Runtime:}
We compare how much time does NER-SQMRC and NER-MQMRC take to do one pass over the complete train data (1 epoch) as well as for inference on complete test data. For a fair comparison, the models are run on the same machine and under the same conditions.

\noindent \textbf{Named Entity Recognition:}
We evaluate models for the task of extracting entities from a given text. For NER-SL models~\citep{Mehta2021LaTeXNumericLT,Ma2016EndtoendSL}, input is a text in which tokens are to be tagged with entity BIO labels (B-PER, I-LOC, etc.). For NER-SQMRC models~\citep{devlin-bert,google-kdd-2020,xu-etal-2019-scaling}, input is a text and a corresponding single entity question, whereas, for our proposed NER-MQMRC models, input is a text and a multi-entity question (section~\ref{problem_def_dataset}). The output for each model (NER-SL, NER-SQMRC and NER-MQMRC) are BIO labels for each token in the text. We use micro average precision (P), recall (R) and F$_1$ as evaluation metrics and use Exact Match criteria~\cite{rajpurkar-etal-2016-squad} to compute the scores.

\noindent \textbf{Few-shot Learning:}
We analyze the performance as the number of data samples seen during training are reduced. We perform this analysis using Ecommerce5PT dataset and compare with Multi-task NER architecture~\citep{Mehta2021LaTeXNumericLT}.


\noindent \textbf{Context-Entity Interaction:}
Element-wise product operation is applied over entity embedding and token output embeddings to get entity specific token embeddings. As the operation performed is important to filter information, in this experiment we explore the effects of using different operations other than using element-wise product.

\noindent \textbf{Impact of entity ordering:}
We evaluate the impact on model performance due to the order in which entities are mentioned in a question as NER-MQMRC is formulated as a multi-entity question.

\begin{table}[]
\centering
\begin{adjustbox}{max width=0.48\textwidth}
\begin{tabular}{@{}lccc@{}}
\toprule
\multicolumn{4}{c}{Ecommerce5PT}                                                                                               \\ 
methods                                                                     & P(\%)          & R(\%)          & F1(\%)         \\ \midrule
\begin{tabular}[c]{@{}l@{}}Multi-task NER~\citep{Mehta2021LaTeXNumericLT} \\ (single model)\end{tabular}     & \textbf{91.62} & 62.47          & 74.29          \\
\begin{tabular}[c]{@{}l@{}}Multi-task NER~\citep{Mehta2021LaTeXNumericLT} \\ (5 model ensemble)$^\ast$\end{tabular} & 88.90          & 77.20          & 82.60          \\
BERT-Tagger~\citep{devlin-bert}                                                                 & 88.43          & 77.51          & 82.61          \\
NER-SQMRC~\citep{devlin-bert}                                                                   & 87.92          & 81.18          & 84.42          \\
NER-MQMRC                                                                   & 87.52          & \textbf{82.14} & \textbf{84.74} \\ \midrule
\multicolumn{4}{c}{AE-Pub}                                                                                                     \\
methods                                                                     & P(\%)          & R(\%)          & F1(\%)         \\ \midrule
SUOpenTag~\citep{xu-etal-2019-scaling}                                                                   & 79.85          & 70.57          & 74.92          \\
AVEQA~\citep{google-kdd-2020}                                                                       & 86.11          & \textbf{83.94} & \textbf{85.01} \\
NER-SQMRC~\citep{devlin-bert}                                                                   & 85.08          & 83.19          & 84.13          \\
NER-MQMRC                                                                   & \textbf{86.18} & 82.97          & 84.54          \\ \midrule
\multicolumn{4}{c}{Twitter}                                                                                                    \\
methods                                                                     & P(\%)          & R(\%)          & F1(\%)         \\ \midrule
BiLSTM-CRF~\citep{Ma2016EndtoendSL}                                                                  & -              & -              & 65.32          \\
CoFEE-MRC~\citep{Xue2020CoarsetoFinePF}                                                                   & 75.89          & 71.93          & 73.86          \\
NER-SQMRC~\citep{devlin-bert}                                                                   & \textbf{80.37} & 76.90          & 78.59          \\
NER-MQMRC                                                                   & 77.79          & \textbf{79.96} & \textbf{78.86} \\ \bottomrule
\end{tabular}
\end{adjustbox}
\begin{tablenotes}
    \item $\ast$ Five individual models were trained and evaluated, one for each product type
    \item AVEQA~\citep{google-kdd-2020} uses no-answer and distillation loss as regularizers
\end{tablenotes}
\caption{Performance comparison on various NER datasets.}
\label{tab:all_datasets_metrics}
\end{table}

\begin{table*}[h]
\centering
\begin{adjustbox}{max width=0.75\textwidth}
\begin{tabular}{@{}llccc@{}}
\toprule
Operation Name       & Formula                         & P(\%)          & R(\%)          & F1(\%)         \\ \midrule
layer\_sum           & $P_i=W_1(T) + W_2(ent_i)$             & 79.07          & 11.46          & 20.02          \\
difference           & $P_i=T - ent_i$                     & 85.99          & 20.70          & 33.36          \\
layer\_product\_relu & $P_i=relu(W_1(T)) * relu(W_2(ent_i))$ & 74.80          & 79.57          & 77.11          \\
layer\_product\_tanh & $P_i=tanh(W_1(T)) * tanh(W_2(ent_i))$ & 76.68          & 78.49          & 77.58          \\
max                  & $P_i=max(T, ent_i)$                 & 76.87          & \textbf{80.79} & 78.78          \\
element-wise product              & $P_i=T * ent_i$                     & 77.79          & 79.96          & 78.86          \\
layer\_product       & $P_i=W_1(T) * W_2(ent_i)$             & \textbf{78.87} & 79.66          & \textbf{79.26} \\ \bottomrule
\end{tabular}
\end{adjustbox}
\begin{tablenotes}
    \item $W_1$, $W_2$ are linear weight matrices
    \item $T$ is the context vector of shape $(n, dim)$ where $n$ is the context length
    \item $ent_i$ is the entity vector of shape $(dim,)$ for the i$^{th}$ entity
    \item $P_i$ is the i$^{th}$ entity specific context vector of shape $(n, dim)$
    \item $*, +, -$ and $max$ are element-wise product, sum, difference and max operations respectively
\end{tablenotes}
\caption{Effect of different operations to attend context vectors using entity vector.}
\label{tab:operations_effect}
\end{table*}

\section{Results}
\label{results}

\subsection{Operational Performance -- training and inference runtime}
\label{sec:latency}
\begin{figure}[htbp]
\centering
\includegraphics[width=0.48\textwidth]{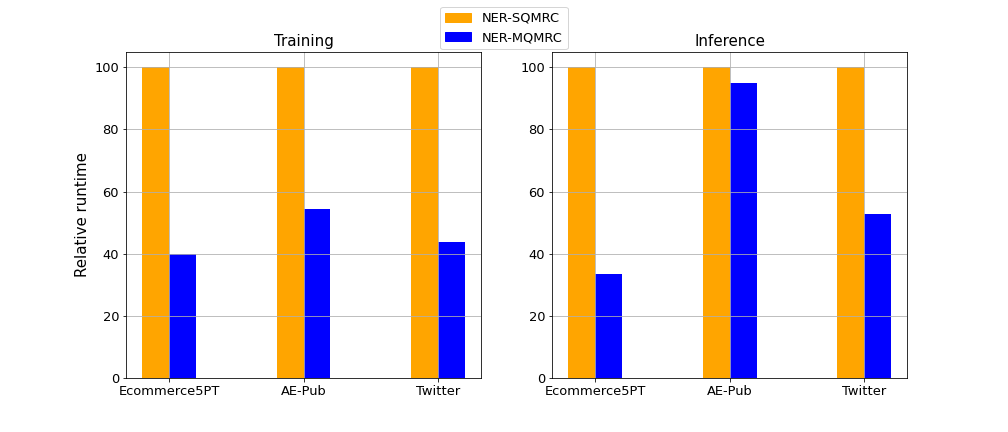}
\caption{Comparison of operational metrics.}
\label{fig:relative_runtime}
\end{figure}

Figure~\ref{fig:relative_runtime} shows the relative training and inference time of NER-MQMRC and NER-SQMRC on all three datasets. We observe that NER-MQMRC leads to an average 2.5 times faster training and 2.3 times faster inference due to performing single forward pass for all entities, as compared to NER-SQMRC which requires a separate forward pass for each entity. The runtime improvement depends on how many entities are grouped together in the dataset for each text. NER-MQMRC inference runtime on AE-Pub is only 5\% faster than NER-SQMRC as only 20.96\% reduction happened in test dataset size after data transformation (Table~\ref{tab:dataset_rows_stats}).

\subsection{NER Task Performance}
Table~\ref{tab:all_datasets_metrics} shows comparison of our proposed model with baselines on multiple NER datasets. Based on evaluation on three NER datasets, our proposed model outperforms NER-SQMRC~\citep{devlin-bert} achieving F1 gain of +0.41\%, +0.32\% and +0.27\% for AE-Pub, Ecommerce5PT and Twitter datasets respectively. A single NER-MQMRC model outperforms ensemble of five Multi-task NER models (one for each product type) by +2.14\% F1 and helps in avoiding model proliferation by having a single model instead of a different model for each product type for Ecommerce5PT dataset. NER-MQMRC outperforms BERT-Tagger~\citep{devlin-bert} by +2.13\% which uses BERT for NER as a tagging task (NER-SL). For AE-Pub, NER-MQMRC has 0.47\% lower F1 compared to AVEQA~\citep{google-kdd-2020}, which is due to the additional No-answer and Distillation loss components in AVEQA. Note that NER-MQMRC is agile and such modules can be easily integrated to it as well.

\begin{figure}[h]
\centering
\includegraphics[width=0.48\textwidth]{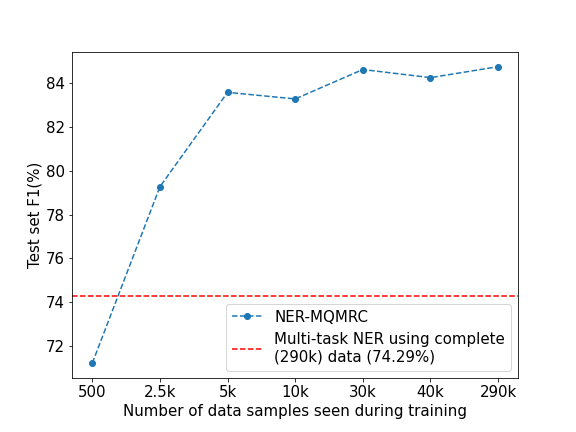}
\caption{Performance with less training data on Ecommerce5PT.}
\label{fig:few_shot}
\end{figure}

\subsection{Evaluation in limited data setting -- Few-shot Learning}
Figure~\ref{fig:few_shot} shows the performance of NER-MQMRC with lesser data availability during training. NER-MQMRC is able to perform better than Multi-Task NER model trained on complete Ecommerce5PT data (290k samples) with as low as 2.5k samples during training. NER-MQMRC is able to perform even with few samples for training because of the natural language understanding a pre-trained BERT model possesses. The performance further increases with increase in dataset size. For Multi-task NER model we observed the F1 further dropped from 74.29\% to 54.49\% when trained with 40k data samples.

\subsection{NER-MQMRC Specific Experiments}
\subsubsection{Context Entity Interaction Operations}
We experimented with a list of different operations to get better entity specific context embeddings on Twitter dataset. As shown in Table~\ref{tab:operations_effect}, \textit{layer\_product} operation performed the best with 79.26\% F1. 
Operations such as element-wise \textit{sum} and \textit{difference} performed poorly in generating good quality entity specific context embeddings because they did not amplify the context vector features by large magnitudes which helps the classification layer better differentiate whereas \textit{product} operation amplified the feature magnitudes.

\subsubsection{Effects of entity ordering in a question}
 We observe that keeping the same ordering of entities in a question while training, leads to deterioration in F1 if the entities are then shuffled during inference (-12.33\% on average). This is likely due to model giving more weightage to relative entity position while learning the entity representations and not focusing on the entity name (or entity question). Shuffling the order of entities during training alleviates this issue and leads to robust results for any order of entities during evaluation.

\section{Industry Impact}
\label{industry_impact}

\noindent \textbf{Cost saving:} Our production pipeline uses AWS p2.8xlarge compute instance for model training which costs \$7.2/hour. Training a single NER-SQMRC model takes 17 hours whereas our proposed NER-MQMRC model takes 7 hours which saves \$72 per model training i.e. reducing the model training cost by an average of 58.82\%. Training multiple such models leads to large cost savings for production systems.

\noindent \textbf{Faster model runtime:} Due to the faster training and inference capabilities of NER-MQMRC, our production systems are deployed faster and are able to serve 2.3 times more inference requests per minute improving the model throughput.

\noindent \textbf{Model proliferation reduction:} The NER-SL based production systems need to deploy multiple models as they are not able to perform at scale with the increase in the number of attributes in the e-commerce catalogue due to the increase in output label space. NER-MQMRC alleviates this issue as the output label space remains constant (3 for BIO labels) and a single model can be trained for 50k+ number of attributes.

\noindent \textbf{Better performance:} From our experiments we show that NER-MQMRC performs better than NER-SL and NER-SQMRC framework models.

\section{Conclusion}
\label{conclusion}
In this paper, we formulated NER as a multi question MRC task (NER-MQMRC). Experimental evaluation on three NER datasets shows that our proposed NER-MQMRC model handles multiple entities together and leads to faster training and inference as compared to single question MRC formulation and improves performance over SOTA NER-SQMRC model~\citep{devlin-bert}, establishing the effectiveness of our proposed model.

\bibliography{output_refs}

\begin{thebibliography}{11}
\expandafter\ifx\csname natexlab\endcsname\relax\def\natexlab#1{#1}\fi

\bibitem[{Devlin et~al.(2019)Devlin, Chang, Lee, and Toutanova}]{devlin-bert}
Jacob Devlin, Ming-Wei Chang, Kenton Lee, and Kristina Toutanova. 2019.
\newblock \href {https://doi.org/10.18653/v1/N19-1423} {{BERT}: Pre-training of
  deep bidirectional transformers for language understanding}.
\newblock In \emph{Proceedings of the 2019 Conference of the North {A}merican
  Chapter of the Association for Computational Linguistics: Human Language
  Technologies, Volume 1 (Long and Short Papers)}, pages 4171--4186,
  Minneapolis, Minnesota. Association for Computational Linguistics.

\bibitem[{Ma and Hovy(2016)}]{Ma2016EndtoendSL}
Xuezhe Ma and Eduard Hovy. 2016.
\newblock \href {https://doi.org/10.18653/v1/P16-1101} {End-to-end sequence
  labeling via bi-directional {LSTM}-{CNN}s-{CRF}}.
\newblock In \emph{Proceedings of the 54th Annual Meeting of the Association
  for Computational Linguistics (Volume 1: Long Papers)}, pages 1064--1074,
  Berlin, Germany. Association for Computational Linguistics.

\bibitem[{Mehta et~al.(2021)Mehta, Oprea, and
  Rasiwasia}]{Mehta2021LaTeXNumericLT}
Kartik Mehta, Ioana Oprea, and Nikhil Rasiwasia. 2021.
\newblock \href {https://doi.org/10.18653/v1/2021.naacl-industry.34}
  {{LATEX}-numeric: Language agnostic text attribute extraction for numeric
  attributes}.
\newblock In \emph{Proceedings of the 2021 Conference of the North American
  Chapter of the Association for Computational Linguistics: Human Language
  Technologies: Industry Papers}, pages 272--279, Online. Association for
  Computational Linguistics.

\bibitem[{Mengge et~al.(2020)Mengge, Yu, Zhang, Liu, Zhang, and
  Wang}]{Xue2020CoarsetoFinePF}
Xue Mengge, Bowen Yu, Zhenyu Zhang, Tingwen Liu, Yue Zhang, and Bin Wang. 2020.
\newblock \href {https://doi.org/10.18653/v1/2020.emnlp-main.514}
  {{C}oarse-to-{F}ine {P}re-training for {N}amed {E}ntity {R}ecognition}.
\newblock In \emph{Proceedings of the 2020 Conference on Empirical Methods in
  Natural Language Processing (EMNLP)}, pages 6345--6354, Online. Association
  for Computational Linguistics.

\bibitem[{Rajpurkar et~al.(2016)Rajpurkar, Zhang, Lopyrev, and
  Liang}]{rajpurkar-etal-2016-squad}
Pranav Rajpurkar, Jian Zhang, Konstantin Lopyrev, and Percy Liang. 2016.
\newblock \href {https://doi.org/10.18653/v1/D16-1264} {{SQ}u{AD}: 100,000+
  questions for machine comprehension of text}.
\newblock In \emph{Proceedings of the 2016 Conference on Empirical Methods in
  Natural Language Processing}, pages 2383--2392, Austin, Texas. Association
  for Computational Linguistics.

\bibitem[{Wang et~al.(2020)Wang, Yang, Kanagal, Sanghai, Sivakumar, Shu, Yu,
  and Elsas}]{google-kdd-2020}
Qifan Wang, Li~Yang, Bhargav Kanagal, Sumit Sanghai, D.~Sivakumar, Bin Shu, Zac
  Yu, and Jon Elsas. 2020.
\newblock \href {https://dl.acm.org/doi/10.1145/3394486.3403047} {Learning to
  extract attribute value from product via question answering: {A} multi-task
  approach}.
\newblock In \emph{{KDD} '20: The 26th {ACM} {SIGKDD} Conference on Knowledge
  Discovery and Data Mining, Virtual Event, CA, USA, August 23-27, 2020}, pages
  47--55. {ACM}.

\bibitem[{Wolf et~al.(2019)Wolf, Debut, Sanh, Chaumond, Delangue, Moi, Cistac,
  Rault, Louf, Funtowicz, and Brew}]{huggingface-transformers}
Thomas Wolf, Lysandre Debut, Victor Sanh, Julien Chaumond, Clement Delangue,
  Anthony Moi, Pierric Cistac, Tim Rault, R'emi Louf, Morgan Funtowicz, and
  Jamie Brew. 2019.
\newblock Huggingface's transformers: State-of-the-art natural language
  processing.
\newblock \emph{ArXiv}, abs/1910.03771.

\bibitem[{Xu et~al.(2019)Xu, Wang, Mao, Jiang, and Lan}]{xu-etal-2019-scaling}
Huimin Xu, Wenting Wang, Xin Mao, Xinyu Jiang, and Man Lan. 2019.
\newblock \href {https://doi.org/10.18653/v1/P19-1514} {Scaling up open tagging
  from tens to thousands: Comprehension empowered attribute value extraction
  from product title}.
\newblock In \emph{Proceedings of the 57th Annual Meeting of the Association
  for Computational Linguistics}, pages 5214--5223, Florence, Italy.
  Association for Computational Linguistics.

\bibitem[{Yu et~al.(2019)Yu, Zhang, Liu, Wang, Li, and Li}]{Bowen2019BeyondWA}
Bowen Yu, Zhenyu Zhang, Tingwen Liu, Bin Wang, Sujian Li, and Quangang Li.
  2019.
\newblock \href {https://doi.org/10.24963/ijcai.2019/750} {Beyond word
  attention: Using segment attention in neural relation extraction}.
\newblock In \emph{Proceedings of the Twenty-Eighth International Joint
  Conference on Artificial Intelligence, {IJCAI} 2019, Macao, China, August
  10-16, 2019}, pages 5401--5407. ijcai.org.

\bibitem[{Zhang et~al.(2018)Zhang, Fu, Liu, and Huang}]{Zhang2018AdaptiveCN}
Qi~Zhang, Jinlan Fu, Xiaoyu Liu, and Xuanjing Huang. 2018.
\newblock \href
  {https://www.aaai.org/ocs/index.php/AAAI/AAAI18/paper/view/16432} {Adaptive
  co-attention network for named entity recognition in tweets}.
\newblock In \emph{Proceedings of the Thirty-Second {AAAI} Conference on
  Artificial Intelligence, (AAAI-18), the 30th innovative Applications of
  Artificial Intelligence (IAAI-18), and the 8th {AAAI} Symposium on
  Educational Advances in Artificial Intelligence (EAAI-18), New Orleans,
  Louisiana, USA, February 2-7, 2018}, pages 5674--5681. {AAAI} Press.

\bibitem[{Zheng et~al.(2018)Zheng, Mukherjee, Dong, and Li}]{opentag}
Guineng Zheng, Subhabrata Mukherjee, Xin~Luna Dong, and Feifei Li. 2018.
\newblock \href {https://doi.org/10.1145/3219819.3219839} {Opentag: Open
  attribute value extraction from product profiles}.
\newblock In \emph{Proceedings of the 24th {ACM} {SIGKDD} International
  Conference on Knowledge Discovery {\&} Data Mining, {KDD} 2018, London, UK,
  August 19-23, 2018}, pages 1049--1058. {ACM}.

\end{thebibliography}
\bibliographystyle{acl_natbib}

\appendix
\newpage
\section{Appendix}
\label{sec:appendix}

\subsection{NER as Single Question MRC}
\label{ner_sqmrc_baseline}
\begin{figure}[ht]
\centering{\includegraphics[width=0.48\textwidth]{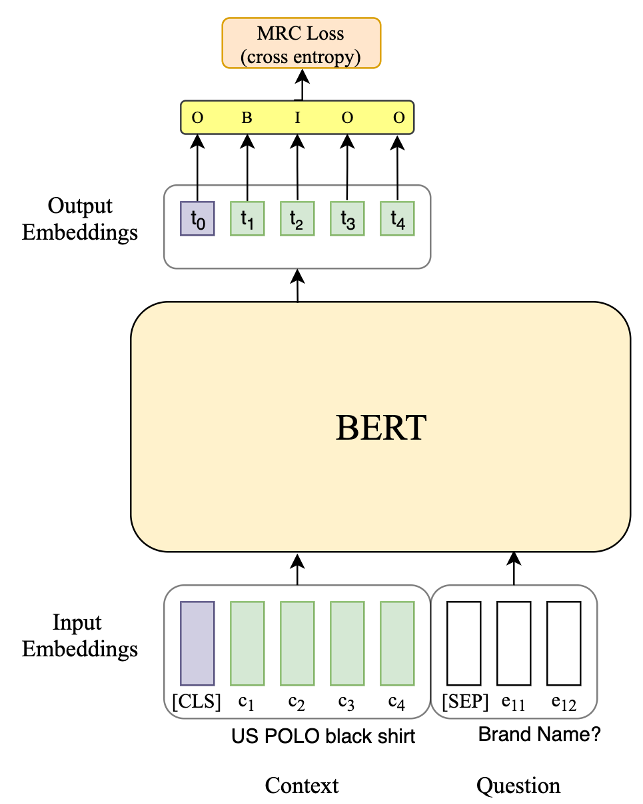}}
\caption{NER-SQMRC model architecture.}
\label{fig:sqmrc_model_arch}
\end{figure}

Figure~\ref{fig:sqmrc_model_arch} shows our baseline NER-SQMRC architecture. We use BERT for Question Answering~\citep{devlin-bert} as our NER-SQMRC baseline implementation. The model is given a question $q_{i}$ asking about $i^{th}$ entity and the model has to extract a text span $x_{start_{i},end_{i}}$ from $X$ which are tokens corresponding to the $i^{th}$ entity. The question component of the input in NER-SQMRC comprises of a single entity of interest to be extracted. The context token embeddings derived from the forward pass of the BERT model are then used to extract the text span corresponding to the entity from the context. For a text with five entities the NER-SQMRC model will need to perform five forward pass through the model to extract the text spans for each of the five entities.

\subsection{Ecommerce5PT training data generation}
\label{ecommerce5pt_data}
Catalogue attribute values can be noisy (e.g. having junk value or missing value) and leads to noisy training annotations with distant supervision. In this section we explain the strategies employed to create better quality training data for Ecommerce5PT dataset.
\subsubsection{Automated Gazetteer}
Using gazetteers in distant supervision can improve the quality of training annotations (especially for attributes which have limited set of valid values). As part of the data tagging step, the catalogue backend values for an attribute are read to create the gazetteer values using the most frequently occuring attribute values. Elbow method is used to determine the threshold for selecting values for the gazetteer. The training data is then created leveraging the backend attribute values and gazetteer values in distant supervision.

\subsubsection{Other Heuristics}
The backend catalogue value sometimes contains a different variation of the attribute value than what is present in the context. For example, context is "US Polo t-shirt for Men" whereas the backend value for the attribute \textit{target-audience} is "Man". Such cases will not be tagged using exact match in distant supervision. Custom heuristics such as pluralizing the text (Men, Mens, Men's, etc.), removing or adding "s", lower casing the text and normalizing attributes such as converting "XXXXL" to "4XL" for \textit{size} attribute are added to improve the training data quality.

\subsection{Implementation Details}
In this section we discuss the dataset creation and model training hyper-parameters details to replicate our results.

During training, we explicitly add no answers for entities that do not have a span in a given text to make the model learn to predict \textit{[CLS]} if no valid answer is present for an entity. We do not make any additions to AE-Pub since it already has no answers added for certain entities. For Ecommerce5PT we add 60\% no answers at random and for Twitter and CoNLL we add all no answers for each entity that is not present in that text.

For our implementation of NER-SQMRC and NER-MQMRC, we use the transformers library \citep{huggingface-transformers}. We use variants of pre-trained BERT model for all our experiments. We use \textit{base-cased} variant for Twitter and \textit{base-uncased} variant for AE-Pub and Ecommerce5PT, keeping our evaluation fairly comparable to existing literature. We use the output layer of single (start, end) span index for AE-Pub dataset similar to~\citep{google-kdd-2020} instead of BIO label. Furthermore, we don't do any dataset specific preprocessing or specific hyperparameter tuning. We use batch size of 32, and a learning rate of 1e-5. We train our models for 20 epochs, choosing the best epoch based on results on the dev set. We make use of AWS compute (ml.p3.8xlarge) instances to run our experiments.

\subsection{Entities per question}
\label{entities_per_que}
\begin{figure}[h]
\centerline{\includegraphics[width=0.48\textwidth]{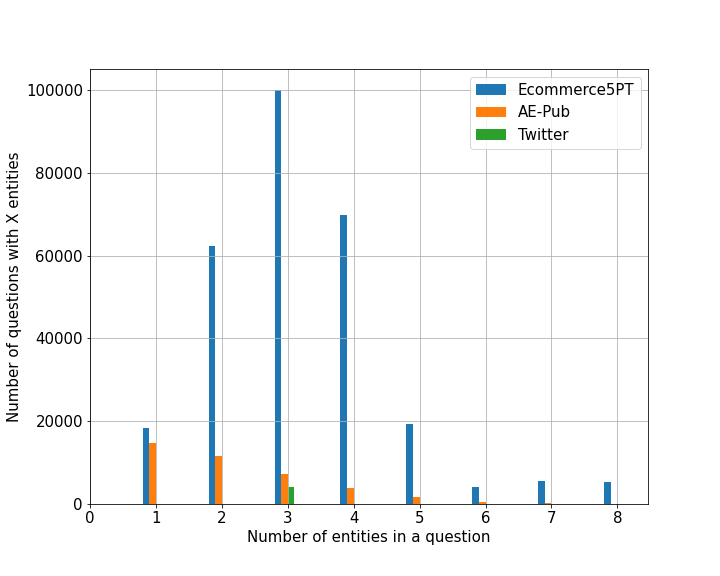}}
\caption{Distribution of number of entities per question in train splits of NER datasets.}
\label{fig:entities_in_que_count}
\end{figure}

Figure~\ref{fig:entities_in_que_count} shows the distribution of number of entities in a question for different datasets. It can be seen from the figure the number of entities in a question greater than 1 are frequent in these datasets which is inefficient for SQMRC type models since they require one forward pass per entity for the same text. We found that Ecommerce5PT and AE-Pub datasets have as many as 12 and 13 attributes for a single text respectively. For Twitter we add all the entities in the question as the dataset has only 3 attributes.

\begin{table}[h]
\centering
\begin{adjustbox}{max width=0.48\textwidth}
\begin{tabular}{@{}cl@{}}
\toprule
Entity Label & Query                                                                           \\ \midrule
PER          & People, persons, including fictional                                            \\
ORG          & Companies, agencies, institutions, organizations                                \\
LOC          & Places, countries, continents, mountain ranges, water bodies                    \\ \bottomrule
\end{tabular}
\end{adjustbox}
\caption{Queries used to replace entity label in a question for Twitter.}
\label{tab:entity_query_map}
\end{table}

\subsection{Queries}
\label{queries}
BERT model has natural language understanding capabilities due to large corpus pre-training. This knowledge can be levaraged in MRC to ask better questions. We use a entity description as question instead of entity name in the question so that better representations can be learned by the model. For Twitter we use the language queries in Table~\ref{tab:entity_query_map}. For Twitter dataset, only \textit{PER}, \textit{ORG} and \textit{LOC} entity label queries are used because we follow the dataset creation guidelines as stated in~\citep{Xue2020CoarsetoFinePF} where \textit{OTHERS} entity label is ignored.


\end{document}